\documentclass[runningheads]{llncs}

\usepackage[T1]{fontenc}
\usepackage{float}
\usepackage{graphicx}
\usepackage[sort]{cite}
\usepackage{booktabs}
\usepackage{hyperref}
\usepackage{color}
\usepackage[absolute,overlay]{textpos}

\hypersetup{
    colorlinks=true,
    linkcolor=blue, 
    urlcolor=blue,
    citecolor=blue,
    pdftitle={SingleStrip},
}

\newcommand{\subpara}[1]{\vspace{0.5mm}\noindent\textbf{#1:}}

\begin{document}

\title{
    SingleStrip: learning skull-stripping \\ from a single labeled example
}
\titlerunning{SingleStrip}

\author{
    Bella Specktor-Fadida\inst{1}
    \and Malte Hoffmann\inst{2-4}
}

\institute{
    Department of Medical Imaging Sciences, University of Haifa, Israel
    \and Athinoula A.~Martinos Center for Biomedical Imaging, Boston, MA, USA
    \and Department of Radiology, Harvard Medical School, Boston, MA, USA
    \and Department of Radiology, Massachusetts General Hospital, Boston, MA, USA
}

\maketitle

\begin{abstract}
    \vspace{-2mm}
    Deep learning segmentation relies heavily on labeled data, but manual labeling is laborious and time-consuming, especially for volumetric images such as brain magnetic resonance imaging (MRI). While recent domain-randomization techniques alleviate the dependency on labeled data by synthesizing diverse training images from label maps, they offer limited anatomical variability when very few label maps are available. Semi-supervised self-training addresses label scarcity by iteratively incorporating model predictions into the training set, enabling networks to learn from unlabeled data. In this work, we combine domain randomization with self-training to train three-dimensional skull-stripping networks using as little as a single labeled example. First, we automatically bin voxel intensities, yielding labels we use to synthesize images for training an initial skull-stripping model. Second, we train a convolutional autoencoder (AE) on the labeled example and use its reconstruction error to assess the quality of brain masks predicted for unlabeled data. Third, we select the top-ranking pseudo-labels to fine-tune the network, achieving skull-stripping performance on out-of-distribution data that approaches models trained with more labeled images. We compare AE-based ranking to consistency-based ranking under test-time augmentation, finding that the AE approach yields a stronger correlation with segmentation accuracy. Our results highlight the potential of combining domain randomization and AE-based quality control to enable effective semi-supervised segmentation from extremely limited labeled data. This strategy may ease the labeling burden that slows progress in studies involving new anatomical structures or emerging imaging techniques.
    
\end{abstract}

\section{Introduction}
Segmentation of anatomical structures in medical images is essential for many clinical tasks, such as assessing tumor size and progression. However, manually labeling is laborious, requires expertise, and results vary with the rater, motivating the development of automated techniques. For example, skull-stripping~\cite{smith2002fast,segonne2004hybrid} of brain magnetic resonance imaging (MRI) is a technique automated in neuroimage analysis pipelines that removes distracting non-brain image content for downstream analyses~\cite{fischl2012freesurfer,jenkinson2012fsl,cox1996afni}. Modern methods leverage deep learning for fast and accurate skull stripping, but they typically rely on high-quality training datasets with annotations, which are costly to create~\cite{tajbakhsh2020embracing}. 
Training strategies using as little data as possible, potentially a single labeled example image, are therefore highly valuable for developing models that segment new anatomies or imaging modalities without substantial labeling effort.

\subpara{Domain randomization}
Recent synthesis, or domain-randomization, strategies~\cite{hoopes2022synthstrip,billot2023synthseg,gopinath2024synthetic,hoffmann2025domain} have shown robust brain segmentation performance across a landscape of modalities by generating diverse training data from a few label maps, which reduces the need for extensive datasets. These methods synthesize images by assigning a random intensity to the voxels of each brain structure and each of background structure. They then apply a series of aggressive augmentation steps, leading to complex intensity patterns across the anatomy. This strategy has led to a number of tools that reliably process unseen image types without retraining. However, despite geometric augmentation, the generated training images do not necessarily cover the real-world anatomical variability needed to train robust, generalizable networks when only a single label map is used.

\subpara{Semi-supervised learning}
Various other approaches address annotation scarcity in semantic segmentation~\cite{tajbakhsh2020embracing}. A prominent example is semi-supervised learning (SSL) with self-training (ST), which creates pseudo-labels. In ST, a model improves itself by first predicting labels for an unlabeled dataset and, second, by adding the pseudo-labels to the training set to learn from the network's own predictions~\cite{cheplygina2019not}. This approach makes it possible to substantially increase the variability of the training data without human effort, as unlabeled datasets are generally more widely available than labeled ones.

\begin{figure}[tb]
    \centering
    \includegraphics[width=11.2cm]{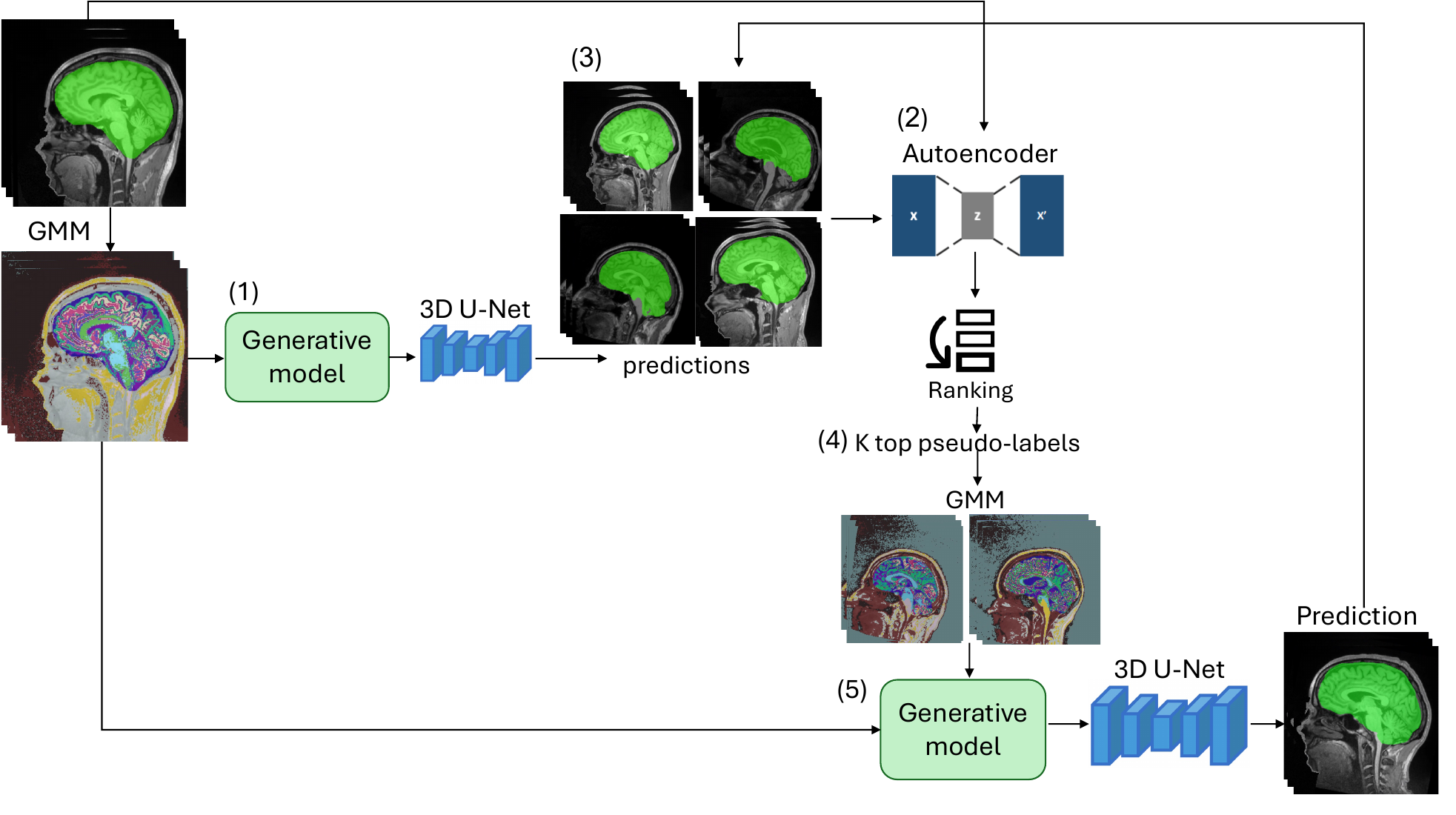}
    \caption{Method overview. To learn skull-stripping from a single labeled image, we first fit a Gaussian mixture model (GMM), assigning voxel intensities to $c$ classes. From these, we synthesize diverse images to train a skull-stripping U-Net. In parallel, we train an autoencoder (AE) to reconstruct brain masks for quality control. Assuming high-quality masks change least in AE reconstruction, we use both networks to skull-strip an unlabeled dataset and retain the least-changing predictions to fine-tune the U-Net via GMM synthesis. This scheme can be repeated several times as needed.\label{fig:method_illustration}}
    \vspace{-2mm}
\end{figure}

Works on ST often train two or more networks in an uncertainty-aware fashion~\cite{tajbakhsh2020embracing,cheplygina2019not,shen2023survey}. One method~\cite{nie2018asdnet} leverages a convolutional adversarial framework combining a segmentation network with a confidence network. The latter predicts an attention map, used to select reliable pseudo-labels predicted from unlabeled data for further semi-supervised learning. Another approach generates pseudo-labels by co-training multiple networks~\cite{xia2020uncertainty,zeng2024segmentation}. It rotates and flips inputs into multiple views and trains separate networks with each, enforcing consistency across unlabeled views weighted by uncertainty. In contrast, consistency regularization~\cite{chen2021semi,ouali2020semi} applies perturbations to generate disagreement across different versions of the same input, training a single model by enforcing prediction consistency for unlabeled data. Recently, an offline ST approach involving teacher and student networks has reached promising segmentation performance~\cite{zoph2020rethinking,specktor2021bootstrap,yang2022st++,specktor2023test}. First, the teacher network trains on labeled data and then predicts pseudo-labels for an unlabeled dataset. Second, the student network trains using both the manual ground-truth labels and the generated pseudo-labels.

\subpara{Quality control}
To reduce the human burden, ST employs automated quality control (QC) techniques to discriminate between reliable and less accurate pseudo-labels. Supervised methods for segmentation QC typically estimate a single metric with a regression network~\cite{robinson2018real,fournel2021medical,huang2016qualitynet,shi2017segmentation,arbelle2019qanet}. However, training a regression network with a single labeled example is challenging. An alternative strategy estimates segmentation quality using a two-dimensional (2D) convolutional autoencoder (AE) trained on a ground-truth label map to reconstruct the shape of the structures of interest from a latent space. A QC metric is then calculated as the Dice score between an input label map and its reconstruction~\cite{galati2021efficient}.

Unsupervised QC techniques aim to gauge segmentation quality in the absence of labels. For example, test-time augmentation (TTA) applies random transformations to a test image and compares the predictions for each version after undoing the transformations. TTA assumes that models are most certain about the predictions with the least variability across transformations and has shown promise in estimating segmentation quality~\cite{Lambert2024-js,specktor2025segqc}.
Recently, a two-stage method~\cite{zhang2025unsupervised} combined realistic tumor synthesis with pseudo-labeling for unsupervised segmentation. Similar to teacher-student approaches~\cite{zoph2020rethinking,specktor2021bootstrap,yang2022st++,specktor2023test}, this method bootstraps a first network and uses its predictions to train a second, focusing on bridging the gap between real and synthetic images.

\subpara{Contribution}
In this work, we combine synthesis techniques and pseudo-labeling with automated quality estimation to bootstrap high-performing segmentation networks in a semi-supervised setting. Instead of aiming for realism, our generative model synthesizes training data far outside the typical range of medical images, prioritizing model exposure to intensity variability. We demonstrate robust performance using as little human input as a single labeled image for training skull-stripping networks.

\begin{figure}
\centering
\includegraphics[width=8cm]{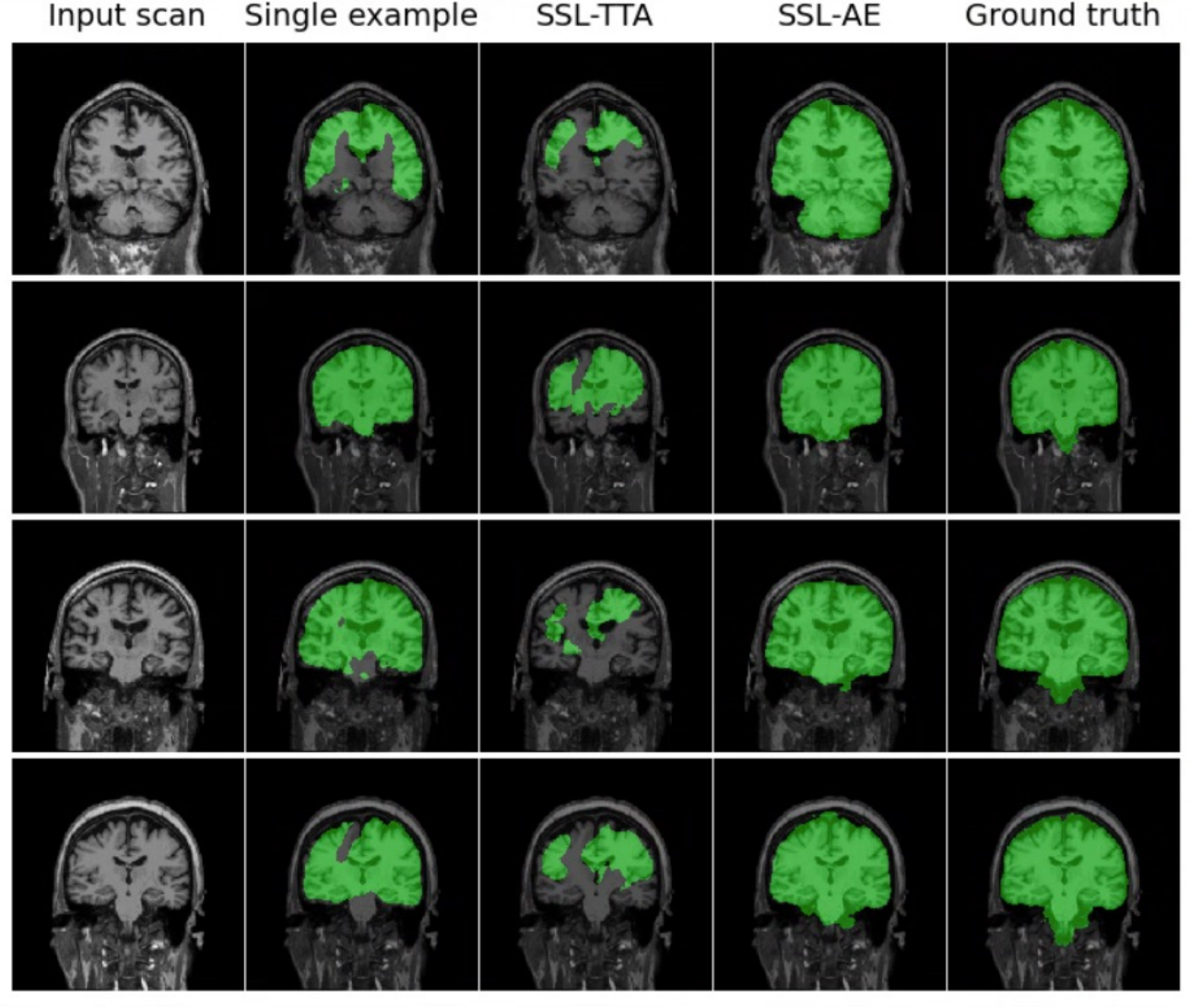}
\caption{Skull-stripping examples. Methods SSL-TTA and SSL-AE fine-tune a U-Net trained with a ``Single example'' using pseudo-labels selected via test-time augmentation and autoencoder reconstruction, respectively. Each row shows a different ASL subject.\label{fig:results_illustration}}
\end{figure}

\section{Methods}
Our semi-supervised approach learns skull-stripping from a single labeled image in five steps (Figure~\ref{fig:method_illustration}). First, we fit a Gaussian mixture model (GMM) to the image, assigning its voxel intensities to $c$ classes. We use these to synthesize a highly variable stream of images and corresponding label maps, to train an initial segmentation network.
Second, we train a convolutional AE to map label maps to a latent space and then reconstruct them---using only the original label map, with augmentation.
Third, we apply the segmentation network to unlabeled data and generate an auxiliary training set with pseudo label maps.
Fourth, we choose the $k$ highest-quality pseudo label maps, based on Dice overlap with their AE reconstructions.
Finally, we fit GMM labels for the auxiliary dataset and retrain the segmentation network on synthetic images generated from the auxiliary training set, combined with the original labeled example. Steps 3--5 can be repeated several times as needed.

\subpara{Domain-randomized skull-stripping}
Building on a recent skull-stripping method~\cite{hoopes2022synthstrip,kelley2024boosting}, we train a three-dimensional (3D) convolutional network to predict brain masks, using synthetic pairs of images and label maps. Briefly, a generative model spatially augments an input label map of $c$ GMM-fitted classes by applying a smooth random deformation including translations, rotations, scaling, and shear. From the new label maps, the model generates an image by assigning a uniformly sampled intensity value to each class in the label map. Finally, a series of randomized corruptions including additive noise, smoothing, cropping, intensity modulation, and exponentiation yield highly complex intensity patterns across the image.

While prior work uses ground-truth label maps with multiple anatomical structures~\cite{hoopes2022synthstrip,kelley2024boosting}, our focus is on bootstrapping a segmentation network with as little human input as possible. Therefore, we start with a single ground-truth brain mask. We separately fit non-anatomical GMM labels inside and outside the brain, and generate input images from them for training. Training optimizes a soft-Dice criterion between the network prediction and the spatially augmented ground-truth brain mask. Rather than developing a new network architecture, we adopt an existing U-Net~\cite{ronneberger2015u,hoopes2022synthstrip}.

\subpara{Autoencoder-based quality control}
To distinguish reliable from less accurate pseudo label maps without human intervention, we employ a reconstruction-based method~\cite{galati2021efficient}. We train a 2D convolutional AE to map an input brain mask to a latent space and reconstruct it, using soft Dice in combination with a mean squared error criterion. Assuming that reconstructions of high-quality masks are accurate while unobserved lower-quality masks lead to inaccurate reconstruction, we consider Dice scores between AE inputs and outputs a proxy for brain-masking quality. We rank 3D brain masks by their volumetric Dice score.

For the AE, we adopt an existing network architecture~\cite{galati2021efficient}, choosing a 100-element latent vector, batch size 8, and learning rate 0.001. We train the AE for 500 epochs with the Adam optimizer, using weight decay $10^{-5}$ and applying rotations between $-30^\circ$ and 30$^\circ$, scaling factors between 0.7 and 1.4, random shift of the center, and flipping for augmentation.

\subpara{Semi-supervised learning}
We employ SSL to improve the initial network using pseudo-labels. Specifically, we apply the trained network to predict brain masks for an unlabeled, auxiliary dataset. We expect these masks to have variable quality, which we evaluate via AE reconstruction. We select the $k$ brain masks with the lowest reconstruction error and add them to the training set. Sampling each pseudo brain mask with the same probability as the initial ground-truth brain mask, we retrain the segmentation network to skull-strip synthetic images generated from $c$ GMM-fitted classes, with fine-tuning of the original single-example (SE) network. We may repeat this step several times, as needed.

\begin{figure}
\centering
\includegraphics[width=10cm]{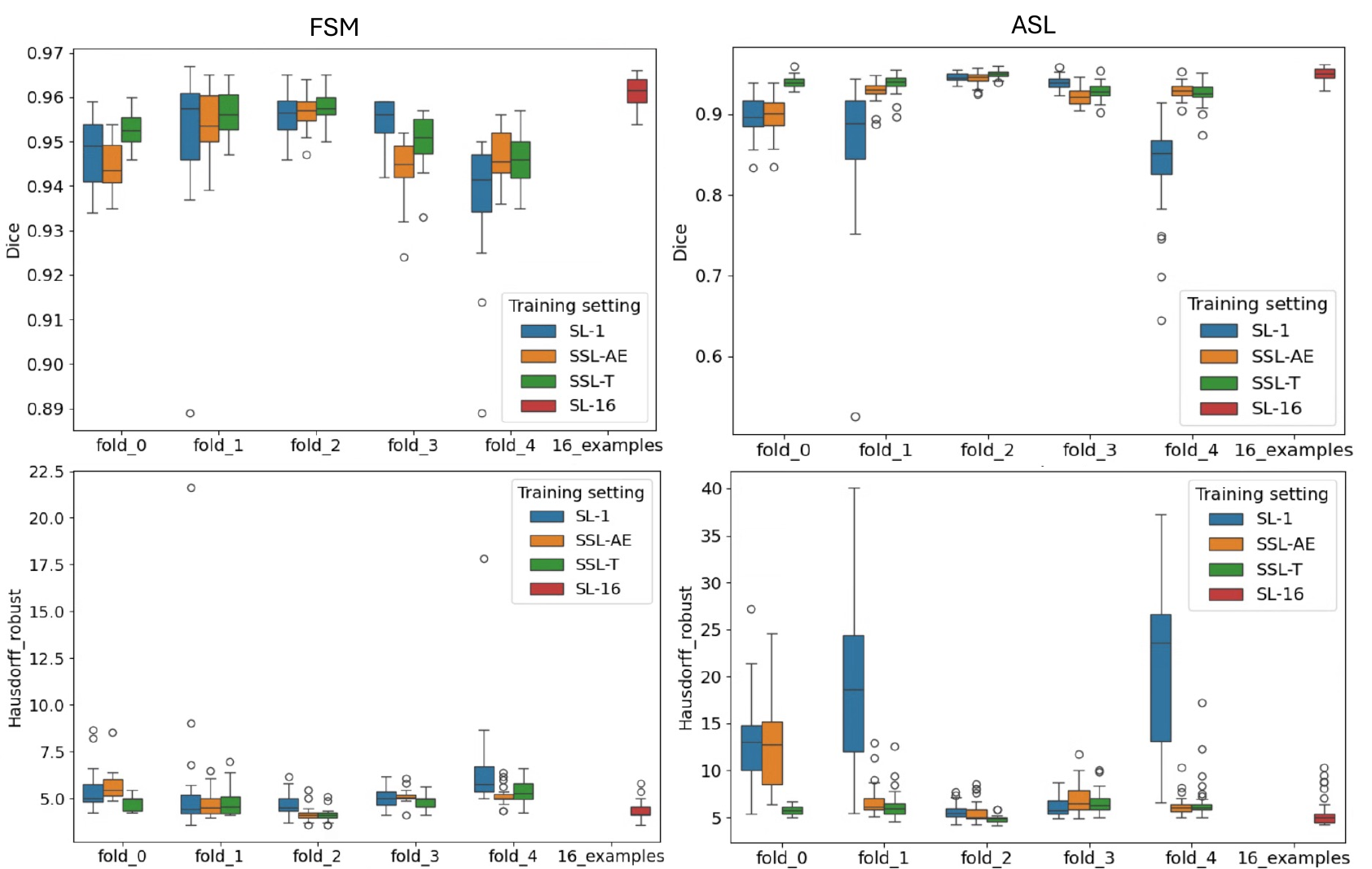}
\caption{Skull-stripping accuracy for in-distribution FSM and out-of-distribution ASL images. We compare training with a single example (SL-1), further fine-tuning using pseudo-labels selected using autoencoder reconstruction (SSL-AE) or pseudo-labels most similar to the ground-truth brain masks (SSL-T), and training with 16 examples (SL-16). Higher Dice and lower Hausdorff values are better.\label{fig:results1}}
\end{figure}

\section{Experiments and results}

\subpara{Data}
We evaluate our SSL strategy at skull-stripping using 81 structural T1-weighted brain-MRI scans with ground-truth brain masks from the SynthStrip dataset~\cite{hoopes2022synthstrip}. We select 38 FreeSurfer maintenance (FSM) images for training and in-distribution (ID) testing, as well as 43 images from the arterial-spin-labeling (ASL) subset for out-of-distribution (OOD) testing. 
We split the FSM subset into training, validation, and test sets of size 16, 2, and 20, respectively. For each image in the training split, we extract $c=16$ GMM classes for image synthesis---10 inside and 6 outside the brain.

\subpara{Setup}
We test 5 folds by training our network on a single, randomly chosen FSM image and associated brain mask, using the remaining 15 FSM and 43 ASL images as the auxiliary, unlabeled dataset.
For each fold, we use the network to derive pseudo-labels for the auxiliary dataset and select the $k=10$ highest-ranking brain masks for fine-tuning.
To assess the utility of AE-based ranking, we compare against TTA-based QC and, as an upper limit, against selecting the predictions that most closely match the ground-truth brain masks. For TTA, we create 10 brain-mask copies by applying random intensity scaling (up to $\pm20\%$) and rotation between $-30^\circ$ and 30$^\circ$, using their voxel-wise median as a pseudo-label.
As a baseline, we also train a network on all 16 labeled FSM examples, and compare it against the proposed strategy using pseudo-labels for fine-tuning vs.\ retraining from scratch.
For each fold, we evaluate performance in terms of Dice scores and 95\textsuperscript{th}-percentile Hausdorff distances (HD95).

\subsection{Results}

\subpara{Semi-supervised learning}
Figure~\ref{fig:results_illustration} shows representative segmentation examples. Figure~\ref{fig:results1} compares training with a single example (SL-1), SSL with pseudo-labels selected using ground-truth masks (SSL-T), and SSL with automatic pseudo-label selection using AE-based QC (SSL-AE). SSL-T slightly improves accuracy on FSM compared to a network trained with a single example from an average Dice score of 0.950 to 0.953 and HD95 from 5.5 to 4.8~mm. For ASL, the difference is more substantial: Dice improves from 0.897 to 0.936 and HD95 from 12.8 to 6.0~mm. We find almost no difference between SL-1 and SSL-AE for FSM---Dice scores increase from 0.949 to 0.950 and HD95 from 5.0 to 5.5 mm, respectively---but performance on the ASL data improves from 0.897 to 0.924 Dice and from 12.8 to 7.0~mm HD95. For most folds, SSL improves skull-stripping accuracy. However, for fold 3 performance compared to SL-1 decreases even when selecting pseudo-labels using the ground-truth masks (SSL-T).

\begin{figure}[t]
\centering
\includegraphics[width=10cm]{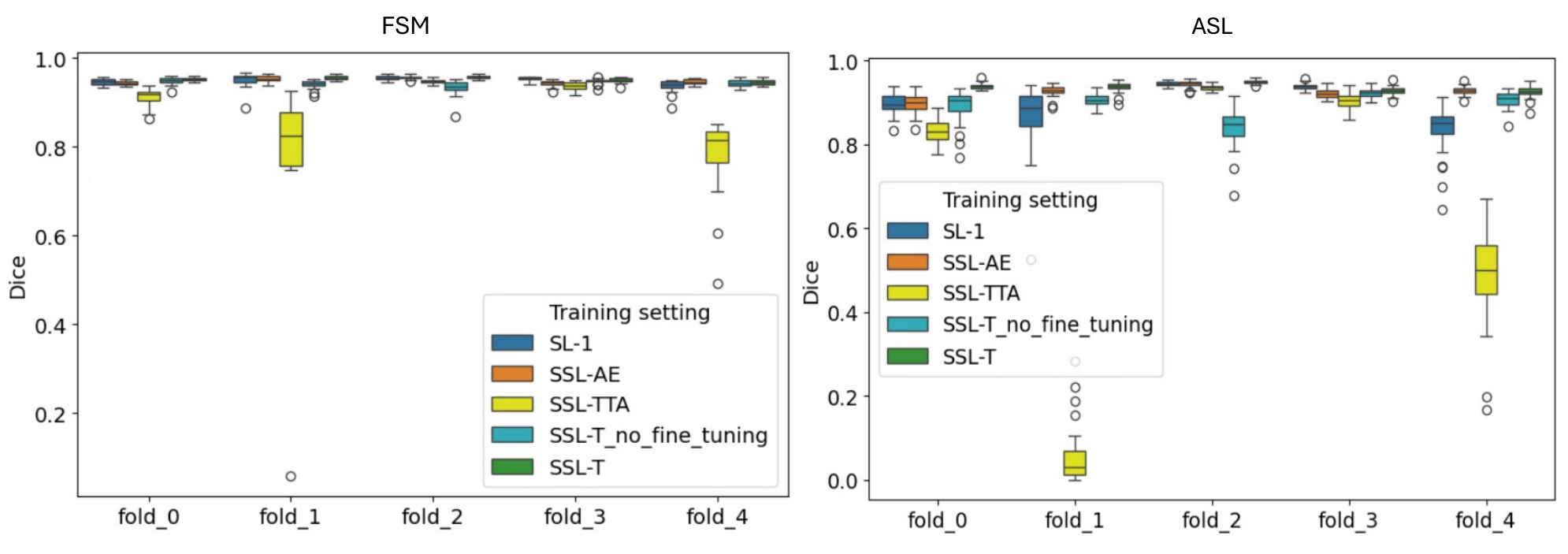}
\caption{Ablation experiments comparing training with a single example (SL-1) to further fine-tuning using pseudo-labels selected with autoencoder reconstruction (SSL-AE), test-time augmentation (SSL-TTA), or via comparison to ground-truth brain masks (SSL-T). In addition, we test retraining from scratch (SSL-T no fine-tuning).\label{fig:results2}}
\end{figure}

Figure~\ref{fig:results2} compares accuracy for our SSL-AE approach against ranking pseudo-labels with TTA (SSL-TTA), as well as using the pseudo-labels for fine-tuning the initial network vs.\ retraining from scratch.
SSL-TTA performs least accurately, with Dice deteriorating from 0.950 to 0.863 for FSL and from 0.897 to 0.624 for ASL, compared to SL-1. Training SSL-T from scratch reduces accuracy compared to fine-tuning, yielding 0.943 vs.\ 0.953 Dice on FSM data and 0.895 vs.\ 0.936 Dice on ASL data.

\begin{figure}[t]
\centering
\includegraphics[width=6cm]{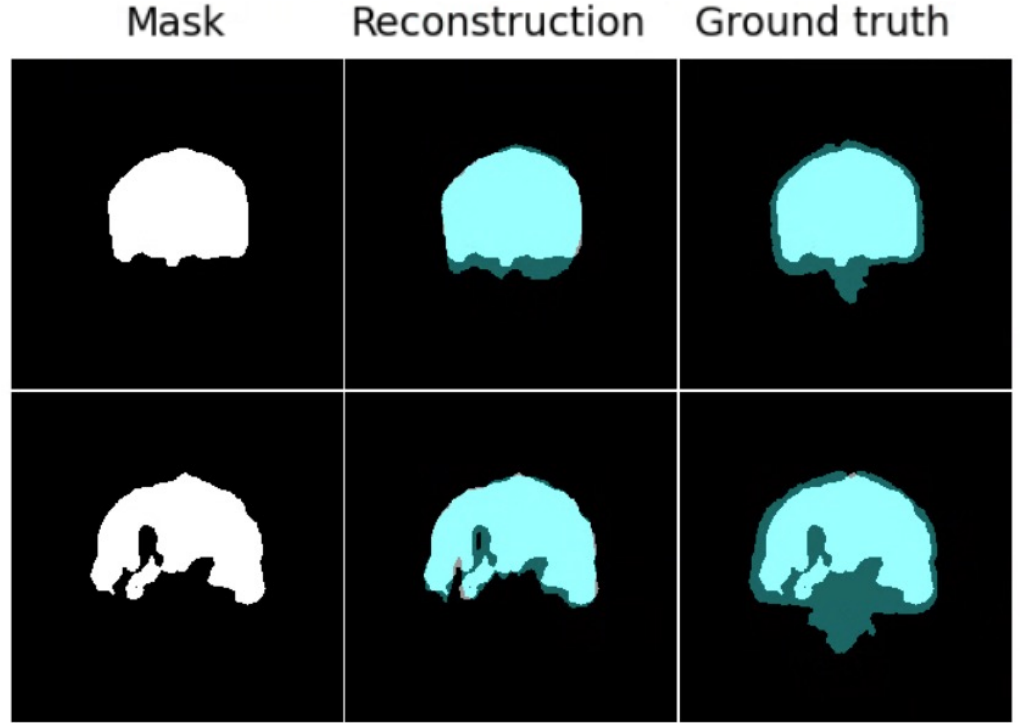}
\caption{Autoencoder (AE) reconstructions. Left: input brain mask. Center: AE output superimposed with the input. Right: ground-truth mask superimposed with the input.\label{fig:reconstruction_results}}
\end{figure}

\subpara{Automated quality estimation}
Table~\ref{table:QC_results} compares Dice scores estimated for pseudo-label selection via AE reconstruction and via TTA against ground-truth scores, in terms of mean average errors (MAE) and Pearson correlation coefficients. For TTA-based QC, we use a model trained on a single label map (SL-1).
While AE-based QC reaches a high average correlation of 0.708 with the ground-truth scores, TTA-based QC yields a weak, negative correlation of $-0.1$, indicating its inadequacy for estimating brain-mask quality. Similarly, the AE achieves a substantially lower MAE of 0.067 than the MAE of 0.107 we find for TTA.
Figure~\ref{fig:reconstruction_results} shows representative AE-reconstructs for a mid-coronal slice. The AE output noticeably differs from the input for both a high-quality and a low-quality brain mask. Yet we find low-quality inputs to undergo more change in reconstruction, such that AE-based ranking of pseudo-labels leads to improved SSL segmentation performance relative to training with a single example.

\setlength{\tabcolsep}{10pt}
\begin{table}
\centering
\caption{Pseudo-label Dice estimation accuracy for test-time augmentation (TTA) and autoencoder reconstruction (AE). We assess agreement with Dice scores between pseudo and ground-truth labels, as mean average error (MAE) and Pearson correlation.\label{table:QC_results}}
\begin{tabular}{ccccc}
    \toprule
            & \multicolumn{2}{c}{MAE $\downarrow$} & \multicolumn{2}{c}{Pearson $\uparrow$} \\
    \cmidrule(l){2-5}
    Fold    & TTA & AE & \phantom{$-$}TTA & AE \\
    \midrule
    0    & 0.079 & \bf 0.048 & $-0.239$             & \bf 0.427 \\
    1    & 0.214 & \bf 0.130 & $-0.170$             & \bf 0.975 \\
    2    & 0.036 & \bf 0.010 & $-0.145$             & \bf 0.771 \\
    3    & 0.043 & \bf 0.025 & \phantom{$-$}$0.111$ & \bf 0.460 \\
    4    & 0.161 & \bf 0.122 & $-0.056$             & \bf 0.909 \\
    Mean & 0.107 & \bf 0.067 & $-0.100$             & \bf 0.708 \\
    \bottomrule
\end{tabular}
\end{table}

\section{Conclusion}
We present a novel SSL framework for training segmentation networks with as little human input as a single labeled example. Our approach combines recent image synthesis techniques with fully automated pseudo-label selection. In a 3D skull-stripping task, the method achieves robust out-of-distribution performance and compares favorably to a network trained on a larger labeled dataset.

Developing a reliable automatic segmentation QC system from a single training example is inherently challenging. In this work, we compare AE-based QC with an unsupervised TTA-based QC method that has shown promise in prior studies. Given only one labeled example, we find TTA-based QC to be unreliable, while AE reconstruction produces pseudo-label rankings that correlate strongly with Dice scores computed against ground-truth labels. This result is somewhat surprising, considering that the 2D AE operates on individual brain-mask slices.

In the future, we will investigate differences between AE reconstructions of low- and high-quality brain masks, and to explore whether a 3D AE can further improve ranking accuracy in the SSL setting. We also plan to evaluate performance scaling with larger training sets and extend our method to more challenging multi-class segmentation tasks.

\bibliographystyle{splncs04}
\bibliography{main}

\end{document}